\documentclass[letterpaper, 10 pt, conference]{ieeeconf}  

\IEEEoverridecommandlockouts                              
\usepackage{graphicx} 
\usepackage{amsmath} 
\usepackage[noadjust]{cite}

\title{\LARGE \bf
	Non-Matrix Tactile Sensors: How Can Be Exploited Their Local Connectivity For Predicting Grasp Stability?
}

\author{Brayan S. Zapata-Impata$^{1,2}$, Pablo Gil$^{1,2}$ and Fernando Torres$^{1,2}$
	\thanks{*Work funded by the Spanish Ministry of Economy, Industry and Competitiveness, European FEDER funds, through the project DPI2015-68087-R (predoctoral grant BES-2016-078290) as well as the project SOE2/P1/F0638 supported by the European Commission's Interreg V Sudoe programme.}
	\thanks{$^{1}$Automatics, Robotics and Artificial Vision Research Group, Dept. of Physics, System Engineering and Signal Theory, University of Alicante, Alicante, 03690, Spain
		{\tt\small \{brayan.impata, pablo.gil, fernando.torres\}@ua.es}}%
	\thanks{$^{2}$Computer Science Research Institute, University of Alicante, Alicante, 03690, Spain}%
}

\begin{document}
	
	\maketitle
	\thispagestyle{empty}
	\pagestyle{empty}

	\begin{abstract}
		
		Tactile sensors supply useful information during the interaction with an object that can be used for assessing the stability of a grasp. Most of the previous works on this topic processed tactile readings as signals by calculating hand-picked features. Some of them have processed these readings as images calculating characteristics on matrix-like sensors. In this work, we explore how non-matrix sensors (sensors with taxels not arranged exactly in a matrix) can be processed as tactile images as well. In addition, we prove that they can be used for predicting grasp stability by training a Convolutional Neural Network (CNN) with them. We captured over 2500 real three-fingered grasps on 41 everyday objects to train a CNN that exploited the local connectivity inherent on the non-matrix tactile sensors, achieving 94.2\% F1-score on predicting stability.
		
	\end{abstract}
	
	\section{INTRODUCTION}
	
	Tactile sensors are being applied in various aspects of the dexterous manipulation process, due to their ability to provide information like surface's properties and acting forces \cite{Kappassov2015}. One aspect under research is the prediction of the grasp stability before lifting a grasped object. Basically, this consists in predicting whether the object will stay in hand or it will slip if we continue to lift it. Predicting the grasp outcome, in this way, is important in many situations because it favours the early detection of grasp failure and the application of corrective strategies throughout the grasp. 
	
	Pressure points or taxels are usually arranged in matrix-like distributions within the surface of a tactile sensor. As a consequence, tactile readings can be interpreted as small images, in which each pixel holds the data from one taxel. This keeps the local connectivity and, therefore, the relationship of the registered values by neighbouring taxels. There are few related works on the processing of tactile readings as if they were images, though they use matrix-like sensors or even internal cameras. 
	
	In \cite{Bekiroglu2010, Bekiroglu2011a, Schill2012, Hyttinen2015}, the authors interpreted tactile readings as images to calculate common image processing features like moments. Then, they trained a Kernel Logistic Regression (KLR) and a Support Vector Machine (SVM) to detect slipping events, adding joint values to enrich the calculated features. In \cite{Meier2016a}, Fourier-related transforms were used to process tactile signals. Afterwards, the authors arranged the resulting vectors in matrices and trained a CNN with them to detect stability states. Although these works treated tactile readings as images, they had to hand-engineer their features.
	
	In contrast, \cite{Cockbum2017} proposed to use autoencoders to autonomously determine those relevant characteristics. They composited a single tactile image by placing side by side the readings of two matrix-like sensors. Then, they built a dictionary of basis vectors using a sparse encoding algorithm. Finally, they trained a SVM to predict grasp stability using the learnt vectors in the dictionary. Similarly, in \cite{Kwiatkowski2017} the authors built a system integrating a composite tactile image from two matrix-like sensors and proprioception of a robot to predict grasp stability with CNNs. These works interpreted tactile readings as images but using matrix-like sensors, so building tactile images was straightforward.
	
	Recently, in \cite{Calandra2017} the authors used a novel tactile sensor with an internal camera, which recorded deformations of the gel inside the sensor throughout its contact with a surface. Hence, the camera captured a picture with the edges in the surface marked against the gel. Using this sensor, they trained a CNN to predict grasp outcomes. In this case, tactile images were real pictures so no re-interpretation was done.
	
	In this work, we interpret as tactile images the readings from three BioTac SP sensors \cite{Syntouch2018}, integrated in a Shadow Dexterous hand. Moreover, we use them to predict grasp stability, transforming this problem into an image classification task. A CNN is implemented and trained in this task since they have proved to be highly performing models for image classification \cite{Krizhevsky2012}. To make it possible, a dataset of more than 2500 grasps of 41 everyday objects was created and it is available at \cite{DatasetRepo}. The main contributions are:
	
	\begin{itemize}
		\item We investigate how non-matrix tactile sensors can be interpreted as images to exploit their local connectivity. Moreover, we demonstrate its application to grasp stability prediction training a CNN. To the best of our knowledge, this is the first work in the literature on learning tactile images from non-matrix sensors.
		
		\item We achieve state-of-the-art performance rates using only a single reading of the tactile sensors. This approach reduces the detection time in comparison with other approaches based on several sequential readings. Compared to other works, we do not need to hand-pick features nor enrich them with proprioceptive data. 
		
	\end{itemize}
	
	\section{METHODOLOGY}
	\label{sec:methodology}
	
	The BioTac SP are a set of tactile sensors developed by SynTouch. They have more electrodes than the previous version and the circuits are integrated in just a single phalanx. It holds 24 electrodes distributed throughout its internal core. These electrodes record signals from 4 emitters and measure the impedance in the fluid located between them and the elastic skin of the sensor. This sensor provides three different sensory modalities: force, pressure and temperature. In this work, we read the force data from three BioTac SP sensors installed on a Shadow Dexterous Hand (index, middle and thumb fingers). 
	
	\subsection{TACTILE IMAGES}
	\label{subsec:tactile_images}
	
	In order to create tactile images from a non-matrix sensor, we propose to design a matrix in which pixels are filled with the electrodes data, but distributed like in the sensor. It is reasonable to believe that keeping their local connectivity is important. Hence, a CNN can learn to recognise features of stability not just on a single electrode's value but also by checking its neighbourhood. 
	
\begin{figure}[h]
		\centering
		\includegraphics[width = 0.475\textwidth]{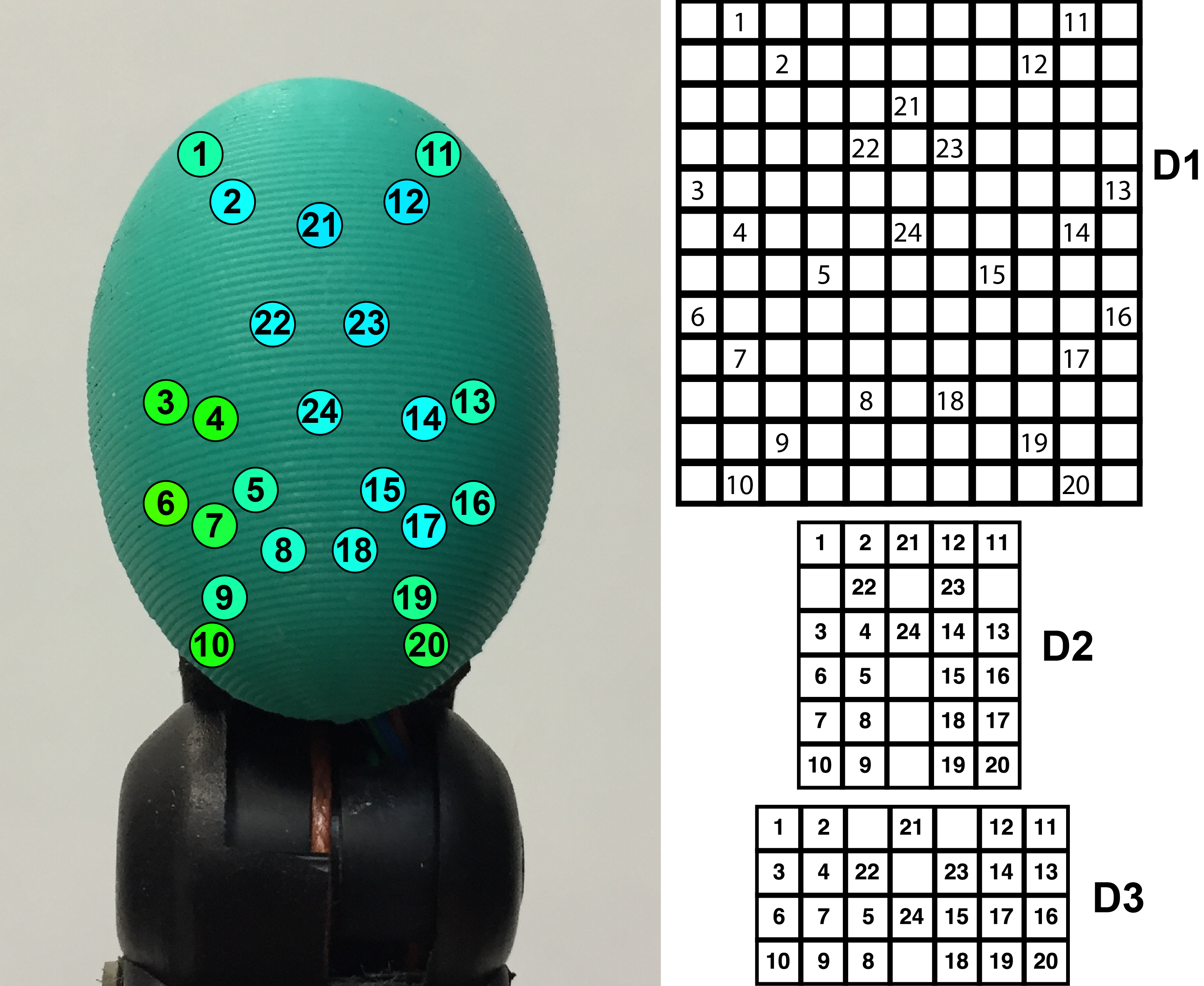}
		\caption{Distribution of the electrodes: (left) on the BioTac SP, (right) on the three proposed tactile images.}
		\label{fig:electrodes-dist}
	\end{figure}

	We present three different distributions (D1, D2, D3) for the BioTac SP sensor, shown in Fig. \ref{fig:electrodes-dist}. We define D1 trying to keep the arrangement of the electrodes so those that are close in the real sensor stay close in the tactile image. Distributions D2 and D3 are an attempt to create a more compact tactile image to avoid gaps in the matrix (blank pixels). The influence of each distribution in the learning process is analysed in section \ref{sec:experimental_results}.
	
	Since we keep the distribution of the electrodes when we build the tactile image, there are pixels that do not correspond to any electrode. The reason for this is that there are already areas in the surface of the sensor without electrodes. For filling these gaps, we have experimented with two strategies, which are discussed in section \ref{sec:experimental_results}: (1) fill these missing values with the value of the less contacted electrode or (2) fill them using the mean of their $k$ neighbours.
	
	\subsection{CONVOLUTIONAL NEURAL NETWORKS}
	\label{subsec:cnn}
	
	We assume that the local connectivity of the electrodes should be useful for predicting grasp stability. When one electrode is recording lower forces, its neighbours should be experiencing similar interactions with the object. Thus, a CNN should be able to exploit this information and perform better than a traditional approach where the electrodes readings are treated as unrelated values.
	
	We calculate a tactile image with the same distribution (D1, D2 or D3) for each finger. We find three ways of preparing them as a single input for a CNN: (1) composite a horizontal image by placing them side by side, (2) composite a vertical image or (3) join them in a three-channel image where each finger corresponds to a channel. We test and discuss these options in section \ref{sec:experimental_results}.
	
	Several CNN architectures were tested, highlighting the three listed in Table \ref{table:cnns} and discussed in section \ref{sec:experimental_results}. We want to note that before training them we normalise the image of each finger separately with the z-score normalisation $z = (x - \mu) / \sigma$, where $z$ is the normalised value, $x$ is the value of a single electrode in the finger, $\mu$ is the mean and $\sigma$ is the standard deviation of the values in the same finger.
	
	\begin{table}[h]
		\renewcommand{\arraystretch}{1.3}
		\caption{Proposed CNN architectures for grasp stability detection.}
		\label{table:cnns}
		\centering
		\begin{tabular}{l|l}
			\hline
			\bfseries Name & \bfseries Architecture                                                                 \\ \hline \hline
			CNN1          & Conv(32x3x3) - FC(1024)                                                     \\
			CNN2          & Conv(32x3x3) - MaxPool(2x2) - FC(1024)                           \\
			CNN3          & Conv(32x3x3) - MaxPool(2x2) - Conv(64x3x3) - FC(1024)       
			\\ \hline
		\end{tabular}
		
		\vspace{3px}
		\scriptsize{Conv($FxKxK$) denotes a convolutional layer with $F$ filters and $KxK$ kernels}\\
		\scriptsize{MaxPool($KxK$) is a Max Pooling layer with $KxK$ kernels}\\
		\scriptsize{FC($N$) denotes a fully connected layer with $N$ neurons}.
	\end{table}
	
	\section{EXPERIMENTAL RESULTS}
	\label{sec:experimental_results}
	
	We have tested the proposed tactile images for grasp stability prediction using 41 objects with distinct geometries, materials, stiffness, sizes and weights. For collecting the training data, we first grasped the object, saved the values of the sensors and later lifted the hand in order to label the stability of the grasp (stable or slippery). We have created two datasets using two orientations for the hand and grasping every object around 60 times in average (Table \ref{table:datasets}).
	
	\begin{table}[b]
		\renewcommand{\arraystretch}{1.3}
		\caption{Datasets created to evaluate our proposal.}
		\label{table:datasets}
		\centering
		\begin{tabular}{lcc}
			\hline
			\bfseries Dataset & \bfseries Stable Grasps & \bfseries Slippery Grasps \\ \hline \hline
			Palm Down           & 667                    & 609                     \\
			Palm Side         & 603                    & 670                     \\ \hline
			Whole            & 1270                   & 1279        \\ \hline           
		\end{tabular}
	\end{table}
	
	To assess our performance, we focus on the F1-score, which is the harmonic mean of Precision and Recall. Every tested model has been evaluated on a 10-fold cross-validation with the \textit{Whole} dataset, unless otherwise specified. The reported results were obtained using a computer with an i7-8750H CPU @ 2.20GHz (6 cores), 16 GiB DDR4 RAM and a GeForce GTX 1060 (1280 CUDA cores) GPU. Everything was coded in Python 3.6, Keras 2.0.8 and TensorFlow 1.8.0.
	
	\begin{figure}[h]
		\centering
		\includegraphics*[width = 0.45\textwidth]{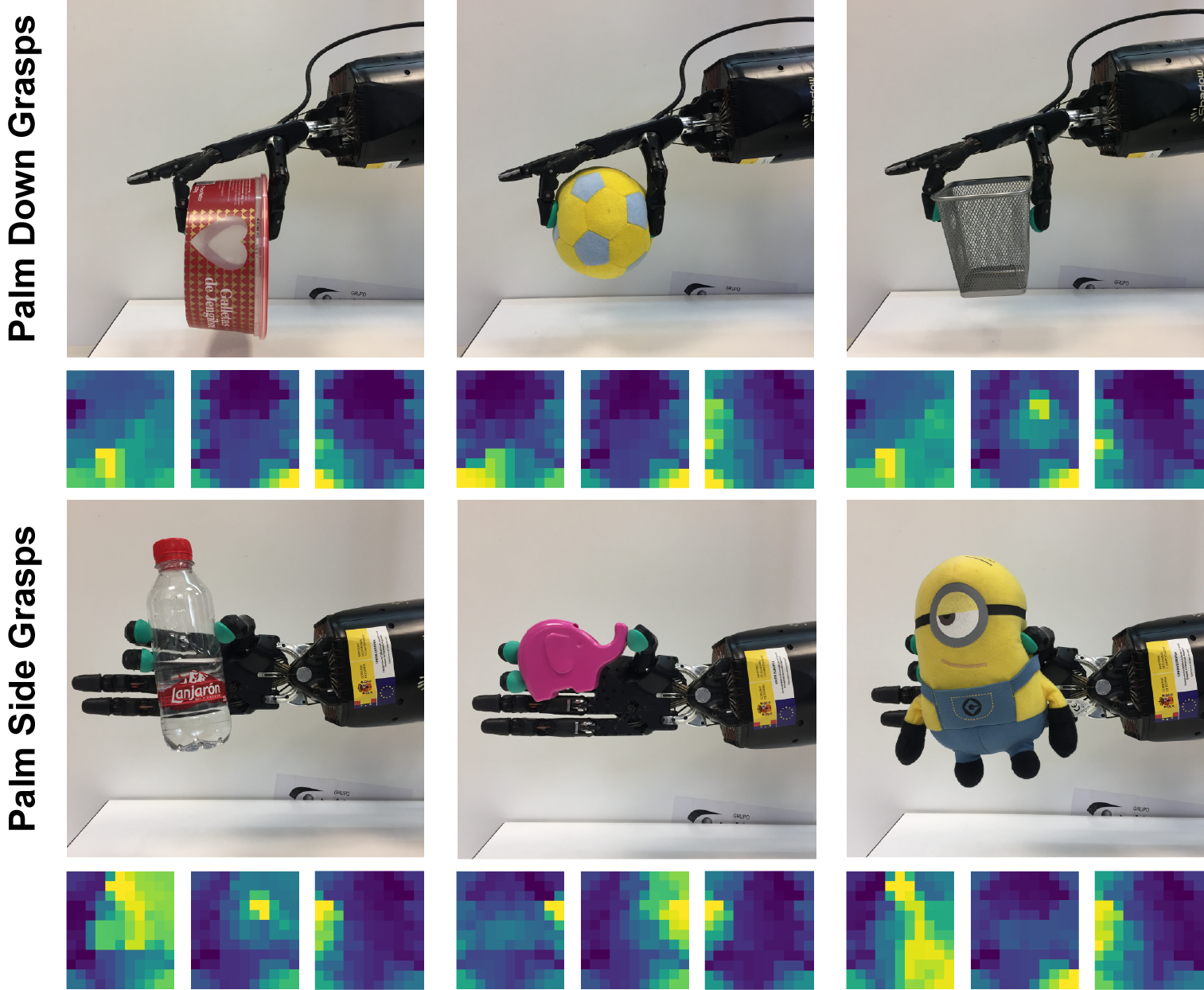}
		\caption{Examples of the proposed dataset. Under each grasping picture, tactile images of the index, middle and thumb fingers are represented using distribution D1 and filling gaps with the mean of the 8-nearest neighbours.}
		\label{fig:grasp-samples}
	\end{figure}                                                                                 
	
	\subsection{TACTILE DISTRIBUTIONS}
	\label{subsec:tactile_distribution}
	
	To check which distribution (D1, D2, D3) worked better, we built a baseline CNN, named as CNN0, with 32 convolutional 3x3 filters followed by a rectified linear unit (ReLU) nonlinearity and fully connected to 128 ReLUs. Finally, this layer was connected to a softmax output layer for the classification of grasps. Table \ref{table:distributions} presents the best results after training CNN0 with each distribution, filling the matrix gaps with the lowest contact value among the electrodes, as introduced in section \ref{subsec:tactile_images}, and joining the matrices to form three-channel images, as explained in section \ref{subsec:cnn}.
	
	\begin{table}[h]
		\renewcommand{\arraystretch}{1.3}
		\caption{Average performance for each distribution.}
		\label{table:distributions}
		\centering
		\begin{tabular}{lcccc}
			\hline
			\bfseries Distribution & \bfseries Acc (\%) & \bfseries Prec (\%) & \bfseries Rec (\%)  & \bfseries F1 (\%) \\ \hline \hline
			D1 (12x11) & \textbf{90.9$\pm$1.5}& \textbf{90.4$\pm$2.4} & \textbf{91.6$\pm$2.4} & \textbf{91.0$\pm$1.4}                         \\
			D2 (6x5) & 89.6$\pm$1.7 & 89.5$\pm$2.8 & 89.9$\pm$2.0 & 89.7$\pm$1.5                         \\
			D3 (4x7) & 89.7$\pm$1.3 & 89.7$\pm$1.8 & 89.8$\pm$2.6 & 89.7$\pm$1.4                         \\ \hline
		\end{tabular}\\
		
		\vspace{3px}
		\scriptsize{ \textit{Acc}: accuracy, \textit{Prec}: precision, \textit{Rec}: recall and \textit{F1}: F1-score.}
	\end{table}
	
	As can be seen, distribution D1 showed the best results, not only on the F1-score but also on the other metrics. This could be due to the fact that D1 has a distribution of the electrodes closer to the real one in the BioTac SP. This shows that arranging them in the tactile image as in the real sensor helps to obtain greater performance rates. In addition, D1 is bigger than the convolutional filters so there are more pixels to work with, something desirable for training this type of networks. Hence, every reported performance on CNN is achieved using D1 since it is our best distribution.
	
	\subsection{GAPS FILLING}
	\label{subsec:gap_filling}
	
\begin{figure}[t]
		\centering
		\includegraphics[width = 0.49\textwidth]{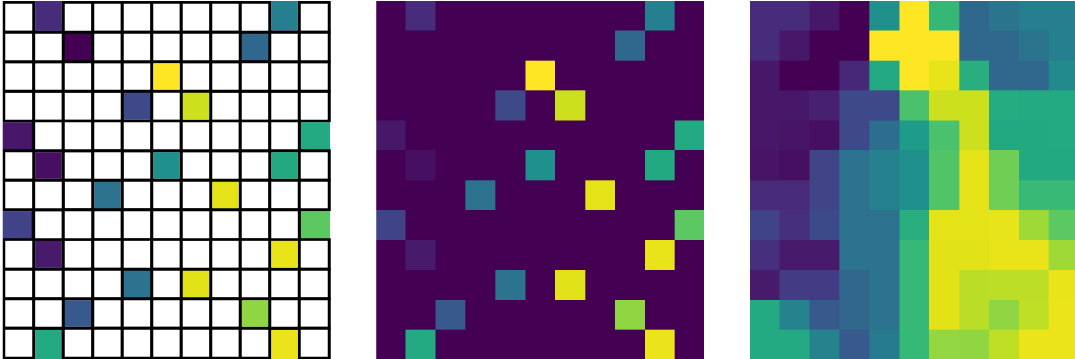}
		\caption{Representation of a tactile image (left) and the result of the two gap filling strategies: (middle) filled with the value of the less contacted electrode and (right) with the mean value of their 8-nearest neighbours.}
		\label{fig:gap-filling}
	\end{figure}

	We experimented with CNN0 to analyse the influence of the gap filling strategies, visually represented in Fig. \ref{fig:gap-filling}, joining fingers in three-channel images, as described in section \ref{subsec:cnn}. When gaps were filled using the value of the less contacted electrode, the model achieved a F1-score of 91.0\%$\pm$1.4\%. For the second strategy, we filled the gaps with the mean value of their 8-closest neighbours using a 3x3 kernel, obtaining a F1-score equal to 92.6\%$\pm$1.7\%.
	
	According to the obtained results, the second method is better than the first one to classify our tactile images. Maybe this occurs because the second strategy smooths the tactile images, helping the network to recognise the patterns that characterise stable and slippery grasps. In contrast, the first strategy creates areas with the same value within the image, fact that could prevent the network from exploiting local connections between electrodes. Therefore, the second strategy for filling the gaps is used in the remaining experiments.
	
	\subsection{IMAGES COMPOSITION}
	\label{subsec:images_composition}
	
	Using the best configuration, CNN0 and D1 with the second filling strategy, we tested three methods for building an input sample: horizontally concatenating the tactile images (12x11) from the 3 fingers to form a 12x33 image (92.1\%$\pm$1.7\% F1-score), stacking them vertically to form a 36x11 image (92.3\%$\pm$1.8\% F1-score) and joining them in depth to form a 12x11x3 image (92.6\%$\pm$1.7\% F1-score).
	
	The last method slightly improved the F1-score, achieving a standard deviation similar to the other strategies. Stacking the images or concatenating them forces the convolutional layers to process the boundaries between the fingers with the same filter. Mixing the values from the fingers like this might have confused the learning of the features during the training process. In contrast, a three-channel image lets CNN0 learn a different filter for each channel. Although these filters are later on combined for joining the channels' values, the obtained performance was better. Following experiments used this strategy for joining the tactile images.
	
	\subsection{LEARNING MODELS}
	\label{subsec:learning_model}
	
\begin{figure}[h]
		\centering
		\includegraphics[width = 0.44\textwidth]{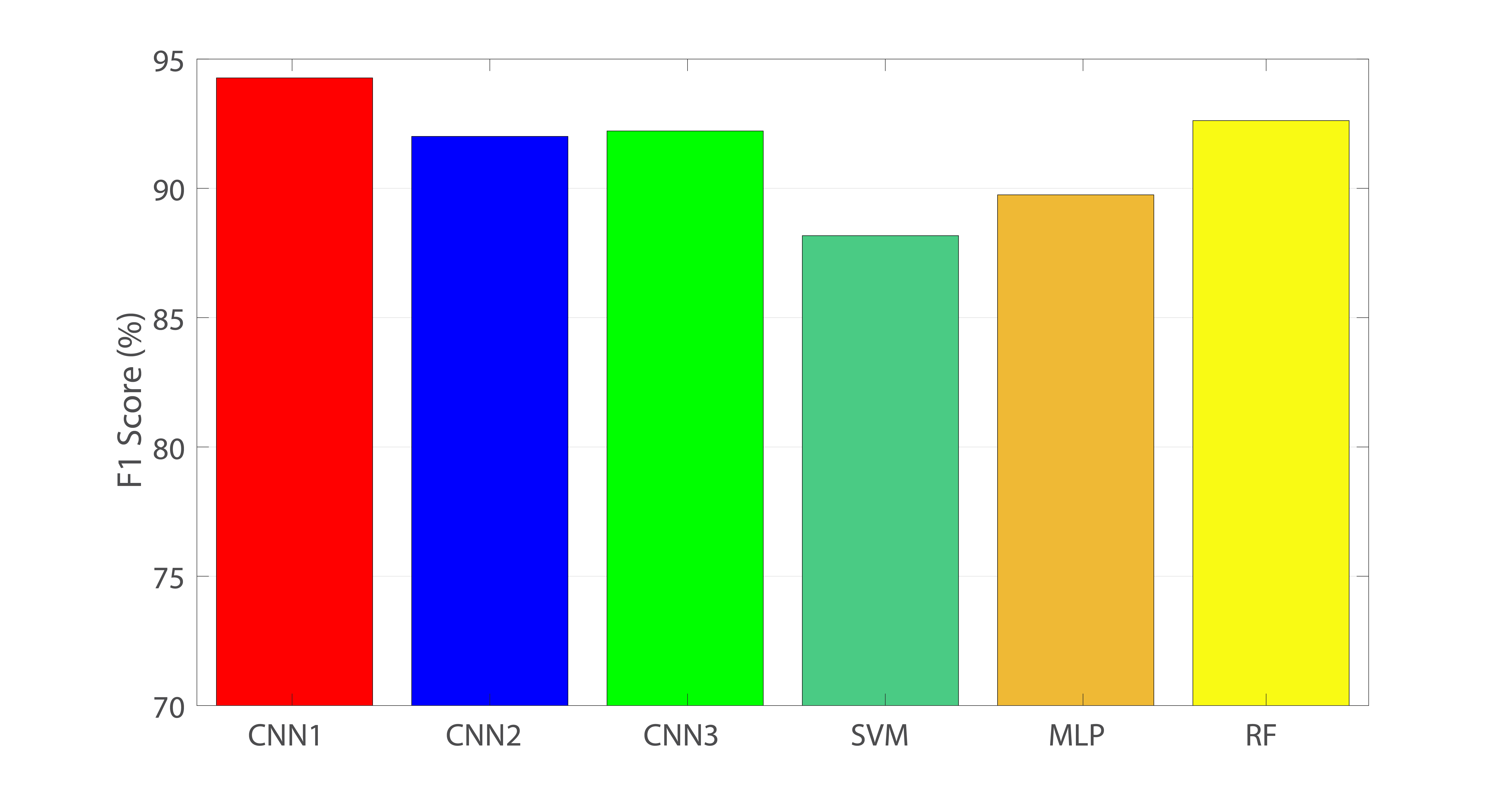}
		\caption{Average F1-score of the tested methods.}
		\label{fig:learning-models}
	\end{figure}

	To prove that tactile images can be exploited efficiently by CNNs on the task of predicting stability, we have compared the performance of the architectures proposed in Table \ref{table:cnns} against a Support Vector Machine (SVM), a Multi-Layer Perceptron (MLP) and a Random Forest (RF). These algorithms are frequently used in machine learning due to their capabilities of generalising a wide range of problems \cite{Caruana2006}. In order to test them, we used the Scikit-learn library to train them with the \textit{Whole} dataset, adapting each grasping sample as a vector with 72 values (24 for each of the 3 fingers). We normalised the vectors before the training phase using the same z-score normalisation applied to the tactile images. Fig. \ref{fig:learning-models} shows the performance achieved by the trained models in terms of F1-score.

	CNN1 achieved the best results (94.2\%$\pm$0.8\%) even though it was the simplest architecture. Adding pooling layers to the architecture reduced the size of the tactile images too much, leading to a lose of valuable information. As a consequence both CNN2 and CNN3 obtained a worse result (92.0\%$\pm$2.0\% and 92.2\%$\pm$2.0\%). Regarding the traditional models, the best result of all of them was obtained by the RF (92.6\%$\pm$1.5\%), being the SVM the worst (88.0\%$\pm$2.9\%).
	
	\subsection{GENERALISATION TESTS}
	\label{subsec:generalisation_tests}
	
	For testing how well this approach would perform on novel objects, we randomly extracted 6 objects from the \textit{Whole} dataset to create a new set called \textit{Unknown}. The remaining 35 objects conformed a dataset called \textit{Known}. This one was used to train CNN1 while the \textit{Unknown} dataset remained for testing. We also trained the RF with the \textit{Known} dataset and tested it using the \textit{Unknown} set to compare its generalisation capabilities with those of the CNN1. We report the average F1-score from 10 training and testing executions, shuffling the samples in the \textit{Known} dataset before each repetition.
	
	The CNN1 performance dropped to 87.7\%$\pm$0.4\% F1-score when it had to predict grasp stability on the \textit{Unknown} dataset. Although the system's performance decreased, it showed a stable behaviour between tests since its standard deviation kept in 0.4\%. Nevertheless, the RF worked better, achieving 90.2\%$\pm$0.5\% F1-score on the same task. This result may be due to the nature of CNNs. They are deep learning models so they require huge amounts of data in order to generalise well. In this experiment, CNN1 was trained using only 2064 samples and tested on 485.
	
	Apart from applying regularisation techniques (dropout, batch normalisation and L2 regularisation) in order to generalise better, we also performed a data augmentation of the training set \textit{Known}. Since we use tactile images, we could be tempted to apply popular techniques for augmenting pictures. However, doing so could create tactile images with improbable values for the electrodes or values misplaced with respect to their real arrangement. Consequently, we performed slight geometrical operations to augment the training set: flipping tactile images vertically, horizontally and rotating them up to $\pm$10 degrees. As a result, the CNN1 trained with the \textit{Augmented Known} dataset (4 times bigger than the original, 8256 samples) increased its F1-score up to 90.1\%$\pm$0.4\%, performing similar to the RF. Therefore, a performance improvement is accomplished with augmentation strategies.
	
	\section{CONCLUSIONS}
	\label{sec:conclusions}
	
	A new trend suggests to interpret tactile readings as images. This has never been tested with non-matrix sensors as far as we know. In this paper, we worked with three BioTac SP tactile sensors for the task of predicting grasp stability using tactile images from non-matrix sensors. We have successfully  proven through extensive experimentation that a CNN can learn to exploit efficiently the local connectivity in these sensors. Since they are deep learning models, regularisation and well-designed augmenting techniques are necessary for boosting their generalisation capabilities.
	
	
	
	\bibliographystyle{IEEEtran}
	\bibliography{tactile}
	
\end{document}